\pdfoutput=1

\documentclass[11pt]{article}

\usepackage[preprint]{acl}

\usepackage{times}
\usepackage{latexsym}
\usepackage{algorithm}
\usepackage{algpseudocode}
\usepackage{amsmath}
\usepackage{graphicx}
\usepackage{subcaption}
\usepackage{fancyvrb}
\usepackage{framed}
\usepackage{tocloft}
\usepackage{listings}
\usepackage{hyperref}
\usepackage{xcolor}
\hypersetup{
    colorlinks=true,
    linkcolor=black,     
    filecolor=black,      
    urlcolor=black,       
    citecolor=black,      
}

\usepackage[T1]{fontenc}

\usepackage[utf8]{inputenc}

\usepackage{microtype}

\usepackage{inconsolata}
\usepackage{tabularx}
\usepackage{graphicx}
\usepackage{subcaption}
\usepackage{booktabs}
\usepackage{multirow}
\usepackage{setspace}
\usepackage{graphicx}

\title{One STEP at a time: Language Agents are Stepwise Planners}

\author{Minh Nguyen\ \ \ \ \ \ \ \ \ Ehsan Shareghi\\
Department of Data Science \& AI, Monash University \\
\texttt{minhtg.nguyen@gmail.com \ \ ehsan.shareghi@monash.edu
}}

\begin{document}


    
\maketitle
\begin{abstract}

Language agents have shown promising adaptability in dynamic environments to perform complex tasks. However, despite the versatile knowledge embedded in large language models, these agents still fall short when it comes to tasks that require planning. 
We introduce STEP, a novel framework designed to efficiently learn from previous experiences to enhance the planning capabilities of language agents in future steps. Concretely, STEP functions through four interconnected components. First, the \textit{Planner} takes on the task, breaks it down into subtasks and provides relevant insights. Then the \textit{Executor} generates action candidates, while the \textit{Evaluator} ensures the actions align with learned rules from previous experiences. Lastly, \textit{Memory} stores experiences to inform future decisions. In the ScienceWorld~\cite{scienceworld} benchmark, our results show that STEP consistently outperforms state-of-the-art models, achieving an overall score of 67.4 and successfully completing 12 out of 18 tasks. These findings highlight STEP's potential as a framework for enhancing planning capabilities in language agents, paving the way for more sophisticated task-solving in dynamic environments.
\footnote{Project page with code: \url{https://github.com/minhtuong201/step.git}}
\end{abstract}

\textit{``You don’t have to see the whole staircase, just take the first step.'' – Martin Luther King Jr. 
}

\section{Introduction}

Autonomous agents that incorporate Large Language Models (LLMs) as integral cognitive systems \cite{coala} have demonstrated significant capabilities in addressing a diverse range of interactive tasks e.g., mathematical problems \cite{gsmk8, math}, programming challenges \cite{bigcodebench, swebench}, and logical reasoning \cite{tafjord2021proofwritergeneratingimplicationsproofs,saparov2023languagemodelsgreedyreasoners}. Nonetheless, their performance tends to diminish in dynamic scenarios, such as Web navigation \cite{zhou2024webarenarealisticwebenvironment, yao2023webshopscalablerealworldweb} and Open-ended environments \cite{scienceworld, alfworld}, which require robust reasoning capabilities of the agents. 

A key contributing factor to language agents' efficiency in long tasks is the notion of memory \cite{coala}. The recent approaches \cite{majumder2023clin, expel} guided the agent to store reflections on their experience of solving a task \cite{reflexion} in memory, and then to retrieve these to improve future attempts \cite{react}. The use of verbal refinements rather than updating model parameters, these techniques are more flexible than conventional Reinforcement Learning (RL) methods. However, this memory module often lacks a retrieval mechanism. Additionally, complex tasks that cannot be solved in a single attempt also require effective planning and goal decomposition mechanisms (we will explain this in more detail later in this paper). 

In this paper, we take a close look at the memory utilization and planning capabilities of the language agent. More concretely, we propose \textbf{STEP} - a novel framework for \underline{Ste}pwise \underline{P}lanning which consists of a \textit{Planner}, an \textit{Executor}, and an \textit{Evaluator}. Upon receiving a task from the environment, the \textit{Planner} decomposes it into manageable subtasks and retrieves relevant information from the \textit{Memory}. After receiving messages, the \textit{Executor} then generates action candidates, which are subsequently evaluated by the \textit{Evaluator} \cite{selfrefine}. Once an action is approved and sent back to the environment, the agent receives an observation and determines whether the subtask requires refinement. After completing an episode, the agent generates learning insights \cite{majumder2023clin}, which are stored in the \textit{Memory} for future attempts. A key aspect is that the \textit{Planner} not only breaks down tasks but also dynamically distils relevant insights from previous attempts to enhance the current task trace. This iterative process maximizes the efficiency of the memory system, ensuring continuous learning and adaptation. 

We evaluate STEP within ScienceWorld \cite{scienceworld} - a dynamic, text-based environment designed to simulate complex scientific tasks. Our results demonstrate that STEP consistently outperforms state-of-the-art (SOTA) models, achieving an overall score of 67.4. The model successfully completes 12 out of 18 tasks, ranking first in 11 tasks.



\section{Background}
\subsection{Reinforcement Learning (RL) Agents}
An RL agent is an autonomous system that learns by interacting with its environment, making decisions, and receiving feedback in the form of rewards to maximize long-term success \cite{ghasemi2024introductionreinforcementlearningfundamental}. We provide a brief review of three works used in our experiment section as baselines:

\textbf{DRRN} (Deep Reinforcement Relevance Network) \cite{drrn} is designed to handle state and action spaces, represented in natural language format. Instead of using raw text, DRRN uses separate neural networks to encode both the game states and possible actions into embedding vectors. Subsequently, it combines them to approximate the optimal action which maximizes the reward. This approach allows the agent to better understand and navigate environments with high-dimensional, unstructured data, showing that it captures meaning instead of memorizing text strings.

\textbf{CALM} (Contextual Action Language Model) \cite{calm} is trained on human gameplay to learn linguistic patterns and common actions, allowing it to generate a set of action candidates for each state. These candidates are then passed to a DRRN agent for re-ranking based on game rewards. By integrating human-like action prediction through a language model with reinforcement learning for assessment, CALM enhances the agent's ability to navigate and interact in complex text-based environments, including unfamiliar situations.

\textbf{KG-A2C} (Knowledge Graph Advantage Actor Critic) \cite{kga2c} enhances the agent’s decision-making by constructing a dynamic knowledge graph during exploration. By leveraging the OpenIE technique \cite{openie}, it maps relationships between objects, locations, and actions, and dynamically updates this graph. KG-A2C helps the agent prune irrelevant actions and focus on the most relevant ones, improving navigation and decision-making in complex environments.

\subsection{Language Agents}
A language agent is an AI system designed for interaction with the external world and understanding natural language, extending the abilities of large language models (LLMs) by incorporating memory mechanisms and related action capabilities~\cite{coala}. 

\subsubsection{Memory}
Memory plays a critical role in language agents, as it enables them to store, retrieve, and process information over time. \citet{coala} categorized memory into short-term and long-term modules including episodic memory, semantic memory and procedural memory. Short-term memory (working memory) holds active information, such as observations from the environment or retrieved knowledge from long-term memory \cite{kang2024thinkactdecisiontransformers}, necessary for immediate processing. As its name implies, working memory serves as a central hub, providing queries to the LLM, translating responses into actionable steps, and linking long-term memory with the current state for cohesive processing.

Episodic memory records the experiences and events the agent has encountered. Like traditional reinforcement learning, where agents adjust their policy based solely on rewards, episodic memory improves learning by enabling more refined understanding and strategic planning, leveraging past successes and failures \cite{ouyang2022traininglanguagemodelsfollow, reflexion, tora}. Semantic memory holds factual information about the world and the agent's identity, capabilities, and task context, acting like a system prompt. For instance, SayCan \cite{saycan} understands it controls a ``physical robot'' to execute real-world tasks described in natural language, while CRITIC \cite{critic} knows its role is to handle free-form question answering or mathematical problems, with the ability to use tools like code interpreters and calculators to verify and refine solutions. Procedural memory comprises the LLM's internal weights and predefined code. Unlike the flexible nature of software-based episodic or semantic memory, procedural memory functions like hardware, requiring precise design to initialize the agent, where any error could lead to bugs or unintended behavior. While some language agents, such as LOGIPT \cite{logipt} and ToRA \cite{tora}, exhibit procedural learning, this usually occurs during fine-tuning phase, not in deployment.

To interact with the memory system, language agents use human-like internal actions - retrieval, learning, and reasoning. Retrieval actions move data from long-term to short-term memory, providing the agent with relevant information to handle tasks. This can be done by directly querying the LLM to retrieve insights based on the current context \cite{expel} or using a key-value method where the key relates to the common in state and the value provides guidance \cite{autoguide}. Learning actions process observations and feedback from working memory, encoding them into long-term memory to enable continuous learning. For example, VOYAGER \cite{voyage}, a Minecraft agent, retrieves skills from episodic memory, generates executable code, and stores new skills gained from interaction with the environment back into long-term memory. Reasoning actions are more complex, involving reading from and writing to working memory. The LLM integrates the task, retrieved insight, probable solutions, and feedback from humans \cite{christiano2023deepreinforcementlearninghuman, ouyang2022traininglanguagemodelsfollow}, self-reflection \cite{selfrefine, reflexion, majumder2023clin} or external tools \cite{critic}. If the solution does not meet the specified constraints, the LLM refines it iteratively.

\subsubsection{State-of-the-art Language Agents}
With the rise of LLMs, there has been a growing body of work around embodied language agents.

\textbf{SayCan} \cite{saycan}, developed by Robotics at Google, integrates LLMs with physical robots to execute real-world tasks based on natural language instructions. Viewing a physical robot as an embodied agent, SayCan itself functions as a language agent, bridging the gap between natural language processing and real-world interaction. While CALM uses DRRN to rerank action candidates, SayCan employs the temporal-difference (TD) RL method to train its value function. This allows SayCan to evaluate the feasibility of suggested actions by accounting for both the current state of the embodied agent and the environment, leading to a more grounded decision-making process. By combining LLMs' semantic knowledge with a contextual understanding of real-world constraints, SayCan efficiently adapts to dynamic environments.

\begin{figure}[t]
    \centering
    \includegraphics[scale=0.4, trim={0cm 0.5cm 0cm 0cm}, clip]{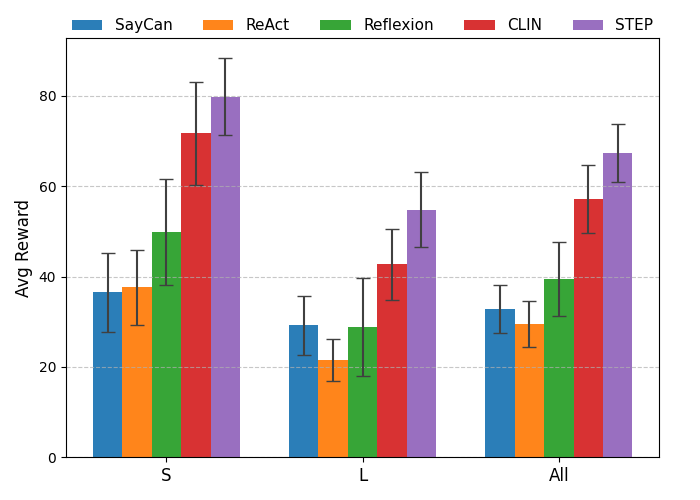} 
    \caption{Comparison of STEP with SOTA.}
    \label{fig:sota}
\end{figure}

\textbf{ReAct} \cite{react} is a simple yet robust technique that integrates reasoning and action generation within language models. Unlike conventional approaches that separate reasoning and action phases, ReAct interleaves these processes, enabling language models to reason before executing actions and adapt dynamically to evolving circumstances. Through the use of reasoning traces, ReAct formulates plans, manages exceptions, and employs actions to acquire additional information from external sources. This integrated approach helps maintain task progress while reducing error propagation and hallucinations. ReAct can apply to a wide range of tasks, including question-answering, fact verification, and other interactive decision-making scenarios.

\textbf{Reflexion} \cite{reflexion} is a framework designed to enhance language agents by enabling self-reflection on past actions and outcomes. Rather than relying on traditional reinforcement learning methods that update model parameters, Reflexion enhances agents through verbal feedback, using linguistic reflection to guide future decisions. While prior models retain feedback only for a single reasoning step \cite{selfrefine, critic, react}, Reflexion stores feedback in an episodic memory buffer, allowing agents to utilize past experiences for more effective decision-making in subsequent tasks. This episodic memory facilitates iterative learning, helping agents avoid repeating errors and continuously refine their strategies. Through iterative self-reflection, Reflexion allows agents to adapt rapidly without the need for extensive retraining, making it a flexible approach for improving performance in complex real-world applications.

\textbf{CLIN} (Continual Learning from Interactions) \cite{majumder2023clin} is a language-based agent framework designed to enhance agents' adaptability by enabling continual learning through dynamic memory updates. Leveraging from Reflexion, CLIN continually refines its memory with \textit{causal abstraction insights} derived from interactions with the environment. This persistent episodic memory enables the agent to reflect on previous trials and incorporate those insights into future decision-making, allowing for rapid adaptation to new environments and tasks without requiring parameter updates. Moreover, with the use of meta-memory, CLIN generalizes its knowledge across multiple tasks and environments, allowing the agent to transfer insights to solve entirely new tasks in different settings. In this paper, we refer to the \textbf{CLIN ADAPT} version, which focuses on rapid adaptation in the same task, same environment.

\subsection{Planning with Language Agents}

Recent research has explored the use of LLMs as planners in various approaches. 

LLM+P \cite{llmp} and LLM-DP \cite{llmdp} utilize LLMs to translate natural language descriptions into the Planning Domain Definition Language (PDDL), enabling classical planners to derive solutions. While symbolic reasoning guarantees finding optimal solutions, it often requires significant human effort for language conversion \cite{huang2024understandingplanningllmagents}, and the assumption of perfect observation of all object states may not hold in dynamic environments. Given the numerous trajectories available to achieve a goal, work on expanding multi-plans in tree structures has been explored. Tree of Thoughts (ToT) \cite{treeofthought}, which employs conventional BFS/DFS for optimal pathfinding, RAP \cite{rap} and LLM-MCTS \cite{llm_mcts} leverage the Monte Carlo Tree Search (MCTS) algorithm for searching. However, these methods are computationally exhaustive, and using LLMs as a world model to evaluate plans becomes impractical, particularly in complex, open-ended environments. 

\begin{figure}[t]
    \centering
    \includegraphics[scale=0.33, trim={0cm 0cm 0cm 1cm}, clip]{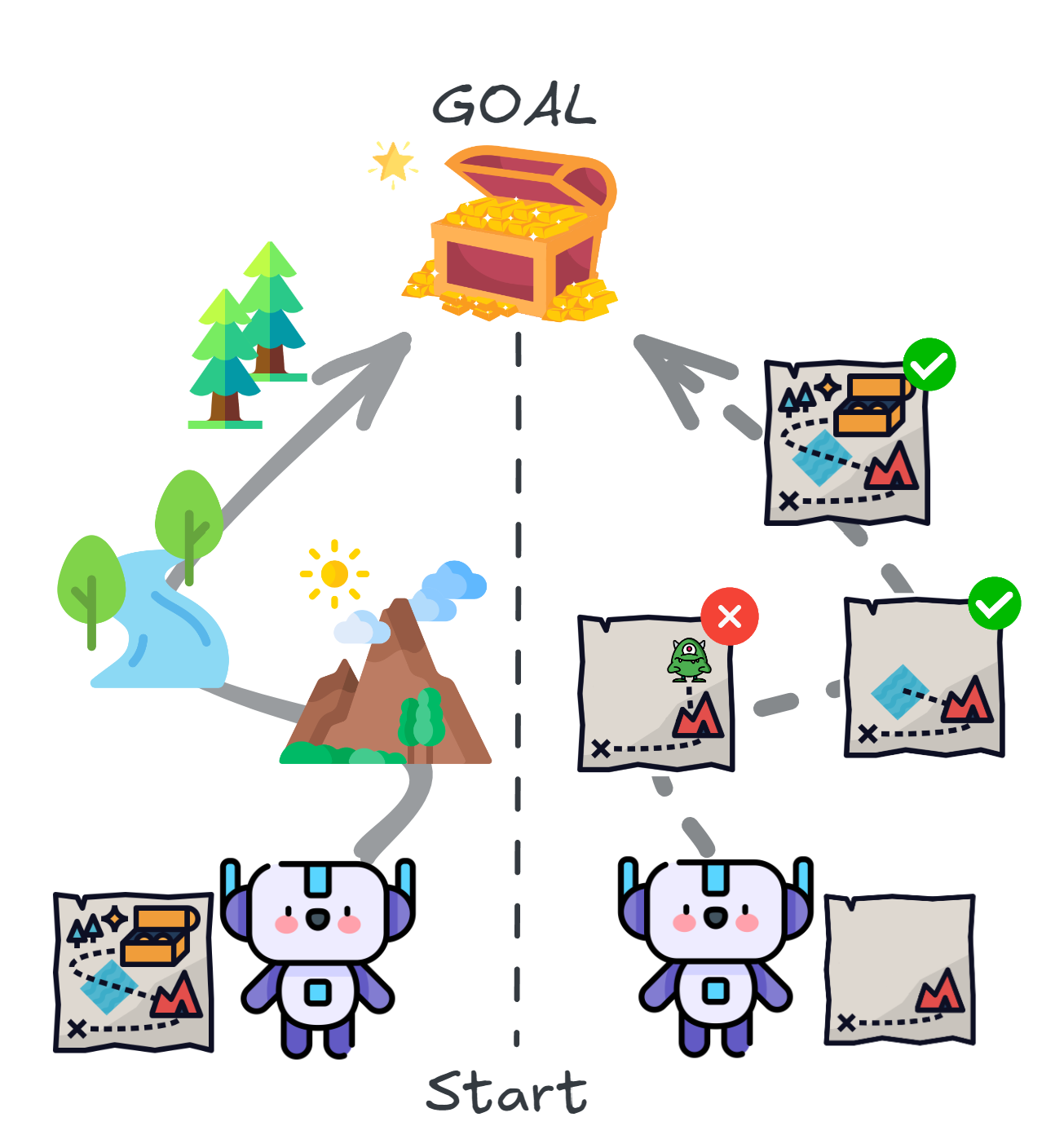} 
   \caption{\textbf{Planning methods}. (LEFT) plan-from-the-start. (RIGHT) plan-on-the-go.}
    \label{fig:plan}
\end{figure}

SOTAs in LLM suggest a different planning method, where instead of tuning the model's parameters \cite{calm, swiftsage}, adjusting the prompts proves advantageous. Two primary schools of thought in planning are ``plan-from-the-start'' and ``plan-on-the-go'' (see Figure \ref{fig:plan}). A representative of the former, Plan-and-Solve \cite{planandsolve}, builds on Zero-shot Chain of Thought (CoT) \cite{kojima2023largelanguagemodelszeroshot} by transitioning from ``Let's think step-by-step'' to ``Let's first make a plan'' and ``Let's carry out the plan''. While this pre-planning strategy offers a straightforward approach of solving task \cite{shen2023hugginggptsolvingaitasks, singh2022progpromptgeneratingsituatedrobot}, it heavily relies on accurate decomposition problems into simpler subproblems at once \cite{zhou2023leasttomostpromptingenablescomplex}. Moreover, predefining a plan before interacting with the environment can result in unrealistic subtasks, requiring adjustments during deployment. 

\begin{figure*}[t]
    \centering
    \includegraphics[scale=0.9]{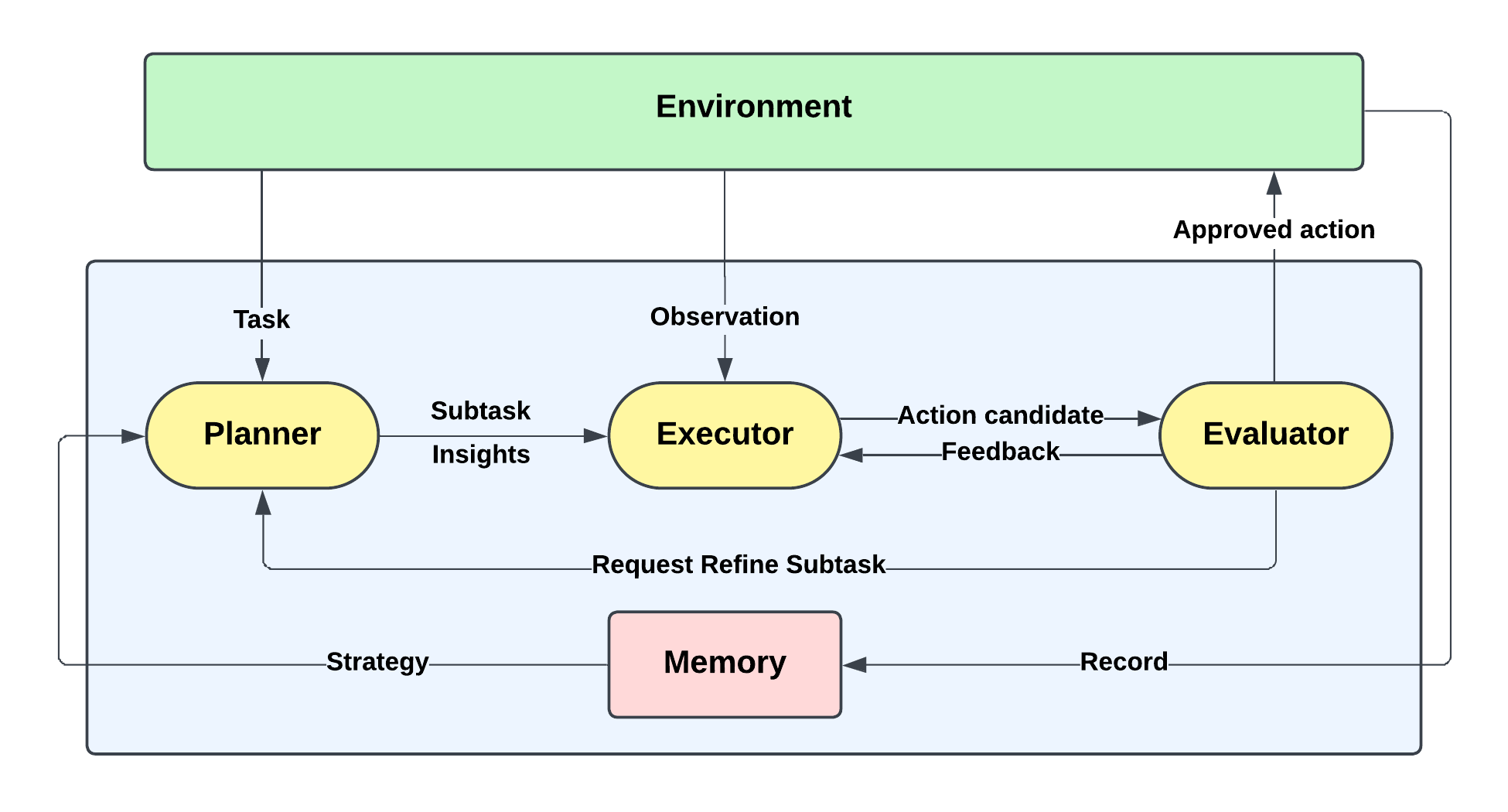}
    \caption{The architecture of \textbf{STEP}. (1) \textit{Planner} receives the task and generates achievable subtasks and relevant insights, (2) \textit{Executor} creates action candidates based on the generated subtasks and insights, (3) \textit{Evaluator} assesses these action candidates for their alignment, and (4) \textit{Memory} stores the experience for future use.}
    \label{fig:framework}
\end{figure*}

In contrast, ``plan-on-the-go'' methods perform better in dynamic environments, where subtasks are revealed one by one during the deploying process. This approach allows agents to improve through feedback-driven retries, without external supervision. ReAct interleaves actions and thoughts for robust planning, while Reflexion adds a verbal reflection after each attempt, showing the self-learning capability of language agents. Building on this, the CLIN model introduces a new type of insight called causal abstractions, which determine whether ``X is necessary for Y''. At each step, the agent accesses a lengthy memory log to determine its next action. While experiments show good adaptability, accessing the entire memory at once may cause the agent to incorrectly prioritize subtasks and risk confusion or overlook key insights. This lack of a transfer learning mechanism \cite{expel} limits its ability to fully leverage learning across different tasks or environments. Building on the CLIN model, we introduce STEP which can maintain task order while leveraging distilled in-context insights from memory, ensuring more accuracy in task execution.

The question of whether language agents can serve as effective planners remains a prominent topic in the Generative AI field. Despite the notable advancements in language agents, \citet{kambhampati2024llmscantplanhelp} argue that LLMs cannot independently perform planning tasks. Their auto-regressive nature limits LLMs from generating fully executable plans or conducting self-verification, which is essential for reasoning and planning. Instead, the authors propose the LLM-Modulo framework, where LLMs contribute by generating candidate plans, while external verifiers refine and validate these plans for accuracy. We agree with this perspective, acknowledging that language agents cannot act as autonomous planners. However, they can serve as stepwise planners, iteratively learning through trial and error. By adapting based on feedback from the external environment, language agents can refine their decision-making process and improve over time, enhancing their planning capabilities within a controlled framework.

\section{Stepwise Planning Language Agent}

We evaluate our model in simulated text-based environments, represented as a partially observable Markov decision process (POMDP). The agents are required to perform sequential actions to accomplish a specific task $M$. Our setup allows an agent to attempt a task multiple times, ranging from \textit{1} to $K$. Each attempt $k$ consists of multiple trials (i.e. episodes), and each trial $T$ comprises a total of $t$ steps. At each step, after issuing an action, the agent receives feedback on environmental changes through observations $o$ and rewards $r$, indicating its performance. After each trial, we enable the agent to make a reflection, generating and storing learning insights in memory for future trials. During the deployment process, STEP consists of three components: \textit{Planner}, \textit{Executor}, and \textit{Evaluator}. At the end of each trial, STEP updates its \textit{Memory} to facilitate continued learning (See Figure \ref{fig:framework}).

\algrenewcommand\algorithmicindent{1em}
\begin{algorithm}
\setstretch{1.1}
\small
\caption{Planning with STEP}\label{alg:step}
\begin{algorithmic}[1]
    \State Task: $M$, Memory: $S, St$
    \For{$k \in 1, \cdots, K$}
        \State Initialize Trial $T$, $t$. $s' = rule(S_{k-1})$
        \While{$t < \text{max. total steps}$ \textbf{or} task not complete}
            \If {$done$}
                \State $m, s = \texttt{Planner}(M, St, S_{k-1}, a_{<t}, o_{<t}, q)$
                \State $q$.append$(m)$
            \Else
                \State $m, s = \texttt{Planner}(M, St, m, S_{k-1}, a_{<t}, o_{<t})$
            \EndIf
            \State $t_0=t$
            \While{$t < \text{max. sub steps} + t_0$ \textbf{or} not $done$}
                \State $g_t, a_t = \texttt{Executor}(m, s, a_{<t}, o_{<t}, f)$
                \State $f, done = \texttt{Evaluator}(m, s', a_{\leq t}, o_{< t})$
                \If {not $f$}
                    \State $r_t, o_t = \texttt{Simulator}(T_{<t}, a_t)$
                    \State $T_{<t+1} = T_{<t} + (r_t, a_t, o_t, g_t)$
                \EndIf
            \EndWhile
        \EndWhile
        \State Final reward $r_k = r_t$
        \State $S_k, St = \texttt{Memory-Generator}(S_{<k}, St, T_k, r_k)$
    \EndFor
\end{algorithmic}
\end{algorithm}

\textbf{Planner} is the strategic component of the framework, responsible for breaking down the current task into manageable subtasks, leveraging a frozen LLM (a model with fixed parameters) to assist in generating the next steps (Algorithm \ref{alg:step}. lines 5-10). At step $t$, the \textit{Planner} receives the main task $M$, the suggested strategy $St$ from the past, and the history of the current trial (a sequence of actions and observations: $\{a_1, o_1, a_2, o_2, \dots, a_t, o_t\}$). Based on the status of the previous subtask $m$, the \textit{Planner} either refines it or derives a new subtask with the list of successful subtasks $q$ to advance the agent toward the main task. Additionally, the \textit{Planner} also retrieves relevant insights $s$ from the learning summary of the previous attempt $S_{k-1}$, which prevents agent access to irrelevant information and focusing on the subtask itself \cite{expel}. 

\textbf{Executor} is the language agent's implementer, tasked with generating appropriate actions to execute the objectives set by the \textit{Planner} (Algorithm \ref{alg:step}. line 13). Following the ReAct concept, the \textit{Executor} is required to derive the next rationale $g_t$ and action $a_t$ given the current subtask $m$ and relevant insight $s$ from the \textit{Planner}. Another modification from the CLIN model is that while the \textit{Executor} is still provided with the sequence of actions and observations, it no longer receives the rationales of previous actions. This is because, at different steps, the \textit{Executor} may have a different subtask $m$, leading to a different motivation for actions. This setup enhances the module's purpose in task performance, while the \textit{Planner} is the sole component managing the overall task flow, the \textit{Executor} has access only to its current subtask. Thus, within the \textit{Executor}, subtasks are converted into concrete actions that can interact with the environment.

\textbf{Evaluator} serves as a quality control mechanism, assessing the action generated by the \textit{Executor} before they are executed in the environment (Algorithm \ref{alg:step}. lines 12-19). While it has been argued that relying solely on internal knowledge for refinement may reduce the performance of language agents \cite{kambhampati2024llmscantplanhelp, critic, valmeekam2023largelanguagemodelsreally}, our \textit{Evaluator} does not select the best action candidate based on its own judgment \cite{selfrefine}. Instead of evaluating the alignment between the action candidate and the task, it focuses on how well the action aligns with the rules $s'$ (e.g. ``\textit{X} does NOT contribute to \textit{Y}'') from $S_{k-1}$. While the dynamic nature of the environment makes it nearly impossible to construct a world model upfront in “plan-from-the-start” approaches, the “plan-on-the-go” strategy takes advantage of this flexibility. After exploring the environment, the \textit{Executor} resembles as simplified world model, which assesses the action candidates. Based on these assessments, the \textit{Evaluator} either sends a feedback $f$ to the \textit{Executor} for refinement or approves it for execution. It also monitors the completion status of subtasks, issuing refinement requests when necessary to avoid exhaustive exploration.

Finally, \textbf{Memory} serves as a critical component in ensuring the agent's continual learning process \cite{tora, reflexion, lats} (Algorithm \ref{alg:step}. line 22). The \textit{Memory} is structured into two primary components: the list of insights $S_k$ (which is inherited from CLIN) and the suggested strategy $St$. At the end of each trial $k$, the agent reflects on its performance $T_{k}$ based on final reward $r_k$, generates new strategy $St$, and updates $S_k$ from previous trials $S_{<k}$. The notion of insights is well-established in recent research and can encompass pairs of actions and observations \cite{calm}, causal verbal reflections \cite{reflexion}, or human feedback \cite{ouyang2022traininglanguagemodelsfollow}. In this work, we adopt the causal abstraction framework proposed by \citet{majumder2023clin}, which evaluates how action $X$ influences action $Y$. This relationship can yield useful insights, such as “$X$ is necessary for $Y$,” or identify errors, such as “$X$ may not contribute to $Y$.”

Although these insights are valuable for guiding the agent’s execution process, a large information may impair retrieval performance, leading the agent to deviate from the intended task sequence. To mitigate this issue and maintain system efficiency, the \textit{Planner} is provided with a suggested strategy $St$ from the most recent trial, enabling it to accurately determine the next subtask and forward relevant insights to the \textit{Executor}.

\section{Experiments}
\subsection{Experimental Setup}

\paragraph{Task Environment.} To assess STEP, we chose ScienceWorld \cite{scienceworld}~\footnote{\url{https://sciworld.apps.allenai.org/explore}}, a text-based interactive environment that simulates elementary science tasks across 10 interconnected locations, including settings like a living room, workshop, and art studio. The environment features 200 object types, such as devices, substances, plants, and animals, and allows for 25 high-level actions (see Appendix~\ref{app:actionspace} for details on action space). When combined with objects, this results in approximately 200,000 possible action-object combinations per step. The environment is dynamically populated with various object arrangements to prevent memorization and promote adaptability. Agents are challenged to perform scientifically reasoned actions, such as testing electrical conductivity or observing biological life stages. Tasks in ScienceWorld vary in complexity, with short tasks requiring fewer than 37 steps to complete, while longer tasks require more than 37 steps.

\paragraph{Evaluation Protocol.} Agents are evaluated based on their ability to complete tasks, with positive scores ranging from 0 to 100. Each task is divided into required steps and optional sub-goals that guide the agent toward the final objective. Performance is primarily measured by how well the agent completes these goals, often requiring it to ``focus on'' the correct object to progress. Critical mistakes, such as focusing on unapproved objects, result in a penalty of -100 and cause the task to reset. In the failure event, the highest positive score achieved before the mistake is recorded to reflect the agent's progress. Otherwise, the final score of each runtime is used, whether the agent successfully completes the task or reaches the maximum step limit per task.

\paragraph{Baselines agents.} In addition to the baseline methods evaluated in the ScienceWorld paper, which included three RL agents - DRRN, KGA2C, and CALM - as well as experiments derived by \citet{swiftsage} involving three generative language agents - SayCan, ReAct, and Reflexion - we also incorporate the CLIN ADAPT version. For further details, please refer to Section 2.1 Reinforcement Learning Agents and Section 2.2 Language Agents.

\paragraph{Configurations.} We use \texttt{gpt-4o-mini} as the base language model for running both CLIN and STEP. Tasks include 9 short tasks (denoted as S) and 9 long tasks (denoted as L). For short tasks, the maximum step limit is set to 37, while for long tasks it is 70. Each task run consists of 5 episodes, where the first episode represents the agent having no prior knowledge of the environment, and the final episode occurs after the agent has learned and updated its memory to adapt to the task. The highest score achieved across these episodes is taken as the score for the current runtime.

\begin{figure}[t]
    \centering
    \includegraphics[scale=0.35, trim={0cm 0cm 0cm 0.5cm}, clip]{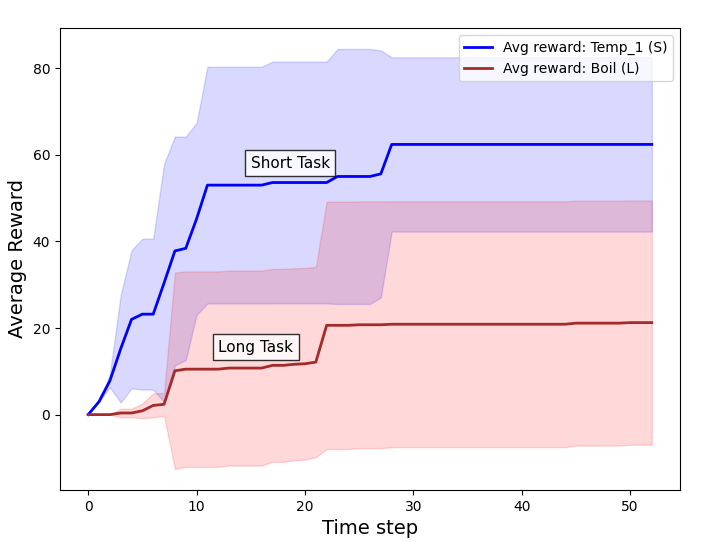}
    \caption{Performance example of STEP in Long and Short Tasks.}
    \label{fig:longshort}
\end{figure}

\subsection{Main Results}

Figure \ref{fig:longshort} presents the model's performance for a short task $\text{Temp}_{1}$ and a long task Boil. The solid lines represent average rewards while the shaded areas capture the variability in performance. Inherited from CLIN, STEP demonstrates strong performance. For the short task, the model shows rapid improvement, reaching an average reward close to 60 within approximately 30 steps, with a noticeable reduction in variability. In contrast, the long task presents slower improvement, with the model's average reward gradually increasing and peaking around 20 after 50 steps, showing higher variability throughout. This suggests that the model adapts more efficiently and stably in the short task while facing greater difficulty and inconsistency in the longer task. See Appendix~\ref{app:trajectory} for a sample from STEP task solving.

Next, we will compare STEP with other state-of-the-art (SOTA) agents, as shown in Figure \ref{fig:sota}. In particular, we will conduct further experiments to compare it with CLIN, the predecessor of STEP, to evaluate performance improvements.

\subsubsection{STEP outperform SOTA Agents}

\begin{table*}[t]
\centering
\resizebox{0.88\textwidth}{!}{%
\begin{tabular}{ll||ccc|cccc|cc}
\toprule
    \multicolumn{2}{l||}{} & \multicolumn{3}{c|}{\textbf{RL Agents}} & \multicolumn{4}{c|}{\textbf{Generative Language Agents}} & \multicolumn{2}{c}{} \\
\cmidrule{0-10}
\textbf{Task} & \textbf{Type} & \textbf{DRRN} & \textbf{KGA2C} & \textbf{CALM} & \textbf{SayCan} & \textbf{ReAct} & \textbf{Reflexion} & \textbf{CLIN*} & \textbf{STEP} \\
\midrule

$\text{Temp}_{1}$
 & S & 6.6 & 6.0 & 1.0 & 26.4 & 7.2 & 5.9 & 9.0 & \textbf{63.0} \\
$\text{Temp}_{2}$
 & S & 5.5 & 11.0 & 1.0 & 8.0 & 6.1 & 28.6 & \textbf{83.3} & 62.7 \\
$\text{Pick\&Place}_{1}$ & S & 15.0 & 18.0 & 10.0 & 22.9 & 26.7 & 64.9 & \textbf{100.0} & \textbf{100.0} \\
$\text{Pick\&Place}_{2}$ & S & 21.7 & 16.0 & 10.0 & 20.9 & 53.3 & 16.4 & 69.3 & \textbf{100.0} \\
$\text{Chemistry}_{1}$ & S & 15.8 & 17.0 & 3.0 & 47.8 & 51.0 & \textbf{70.4} & 55.3 & 61.0 \\
$\text{Chemistry}_{2}$ & S & 26.7 & 19.0 & 6.0 & 39.3 & 58.9 & 70.7 & \textbf{100.0} & \textbf{100.0} \\
$\text{Lifespan}_{1}$ & S & 50.0 & 43.0 & 6.0 & 80.0 & 60.0 & \textbf{100.0} & \textbf{100.0} & \textbf{100.0} \\
$\text{Lifespan}_{2}$ & S & 50.0 & 32.0 & 10.0 & 67.5 & 67.5 & 84.4 & \textbf{100.0} & \textbf{100.0} \\
$\text{Biology}_{1}$ & S & 8.0 & 10.0 & 0.0 & 16.0 & 8.0 & 8.0 & 28.0 & \textbf{32.2} \\
Boil & L & 3.5 & 0.0 & 0.0 & \textbf{33.1} & 3.5 & 4.2 & 4.0 & 21.5 \\
Freeze & L & 0.0 & 4.0 & 0.0 & 3.9 & 7.8 & 7.8 & 32.3 & \textbf{50.9} \\
GrowPlant & L & 8.0 & 6.0 & 2.0 & 9.9 & 9.1 & 7.3 & 30.3 & \textbf{71.5} \\
GrowFruit & L & 14.3 & 11.0 & 4.0 & 13.9 & 18.6 & 13.0 & \textbf{19.3} & 14.0 \\
$\text{Biology}_{2}$ & L & 21.0 & 5.0 & 4.0 & 20.9 & 27.7 & 2.6 & \textbf{59.3} & 46.5 \\
Force & L & 10.0 & 4.0 & 0.0 & 21.9 & 40.5 & 50.6 & 73.3 & \textbf{80.0} \\
Friction & L & 10.0 & 4.0 & 3.0 & 32.3 & 44.0 & \textbf{100.0} & 56.7 & 73.3 \\
$\text{Genetics}_{1}$ & L & 16.8 & 11.0 & 2.0 & 67.5 & 25.7 & 50.9 & 69.8 & \textbf{84.2} \\
$\text{Genetics}_{2}$ & L & 17.0 & 11.0 & 2.0 & \textbf{59.5} & 16.8 & 23.7 & 39.0 & 51.8 \\
\midrule
 & {\textbf{S}} & 22.1 & 19.1 & 5.2 & 36.5 & 37.6 & 49.9 & 71.7 & \textbf{79.9} & \\
 & {\textbf{L}} & 11.2 & 6.2 & 1.9 & 29.2 & 21.5 & 29.2 & 42.7 & \textbf{54.9} & \\
 & {\textbf{All}} & 16.7 & 12.7 & 3.6 & 32.9 & 29.6 & 39.4 & 57.2 & \textbf{67.4} & \\
\bottomrule
\end{tabular}
}
\caption{Comparison of STEP with baselines in ScienceWorld environment. *We use the same backbone LLM, \texttt{gpt-4o-mini}, for both CLIN and STEP experiments for fairness. Type S and L denote short and long tasks, respectively.}
\label{tab:comparison}
\end{table*}

Table \ref{tab:comparison} compares the performance of various agents across 18 tasks in the ScienceWorld environment. The results highlight the superior performance of LLM-based methods (Generative Language Agents) over conventional RL agents due to their advanced generalization abilities, though they come with higher deployment costs. Among the RL agents, DRRN achieves the highest overall score of 16.7, which is still significantly lower than the scores of all the generative language agents, with the lowest among them being ReAct at 29.6. Reflexion demonstrates a strong capability by utilizing episodic memory to retain learning insights across trials, showing more than a 10-point improvement over SayCan and ReAct in short tasks. Building upon Reflexion's memory-based advantages, CLIN benefits from causal abstraction insights, which significantly boosts its performance to a 57.2 overall score, demonstrating its effectiveness in both task types.

Notably, STEP consistently outperforms all SOTA models across both short and long tasks, achieving 67.4 points overall. In short tasks, STEP achieves an impressive 79.9, representing an 11.4\% increase over its predecessor. In long tasks, STEP further demonstrates its ability to handle more complex challenges, with a score of 54.9, marking a 28.6\% improvement over CLIN. Analyzing individual tasks, STEP ranks first in 11 out of 18 tasks, clearly showcasing its dominant performance. For example, in the short task $\text{Temp}_{1}$, STEP significantly boosts the second highest score from SayCan’s 26.4 to an impressive 63.0. In the long task GrowPlant, it doubles the score of 30.3 in CLIN to 71.5. This substantial leap in performance across individual tasks emphasizes STEP's capacity to generalize well and adapt to a wide range of environments.

\subsubsection{STEP outperforms CLIN in continual learning}

\begin{figure*}[t]
    \centering
    \includegraphics[trim= 0 0cm 0 0,clip, scale=0.36]{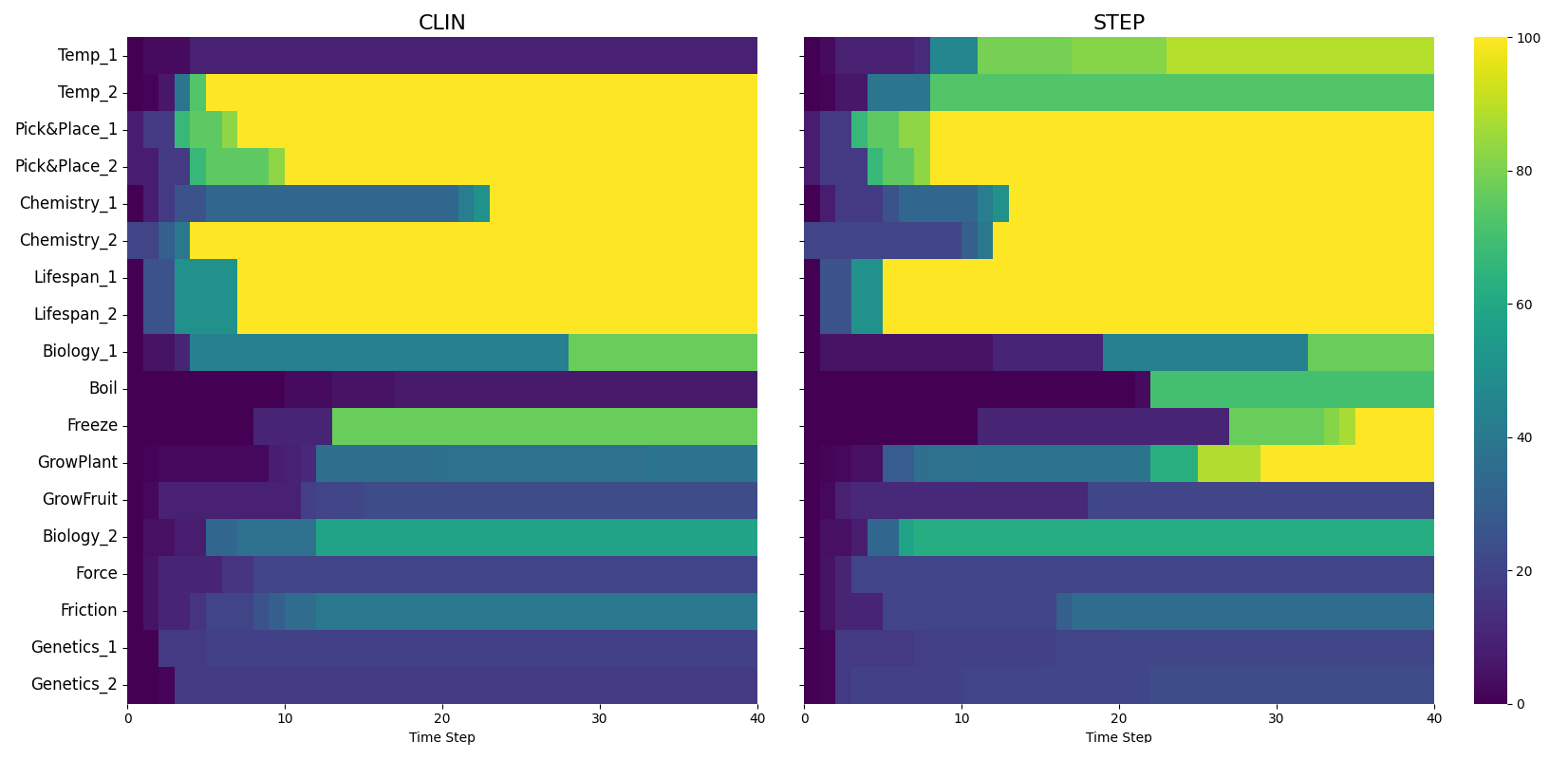} 
    \caption{Task performance comparison between STEP and CLIN.}
    \label{fig:heatmap}
\end{figure*}

Next, we compare the performance between STEP and CLIN across all tasks (see Figure \ref{fig:heatmap}), reporting the best traces while excluding false-positive cases where the agent unintentionally ``cheats'' the task (e.g. in the Friction task, the agent might randomly focus on an inclined plane and complete the task by chance, without adhering to the intended experimental procedure). In short tasks, both CLIN and STEP exhibit strong performance, as indicated by the prominent yellow regions for tasks such as Pick\&Place, Chemistry and Lifespan. However, a key difference arises in $\text{Temp}_{1}$, where CLIN struggles, achieving only 10 points overall. In contrast, STEP shows a gradual progression, steadily achieving subgoals and eventually reaching nearly 90 points. This demonstrates STEP’s capability to break down tasks and make incremental progress, where CLIN appears to fall short.

The performance gap widens further in long tasks, where STEP consistently outperforms CLIN. CLIN shows difficulties in tasks such as Boil, Freeze, and GrowPlant, marked by darker shades in the heatmap, indicating poorer performance. STEP, on the other hand, demonstrates significant improvements in these complex tasks, with more yellow regions, suggesting higher scores and better task completion. A potential explanation for this distinction is that CLIN completes all tasks within 30 steps, as shown by the cutoff in its heatmap, suggesting faster but possibly incomplete learning. Conversely, STEP continues learning beyond that, which allows it to achieve superior overall performance and higher scores, particularly in tasks where CLIN has plateaued. For example, in Freeze, while CLIN initially learns quickly, achieving around 80 points within the first 14 steps, it subsequently halts progress or becomes stuck in a loop. In contrast, after experiencing stagnation up to 15 steps without any improvement, STEP rapidly peaks at the $\text{27}^{th}$ and successfully completes the task at the $\text{36}^{th}$ step. This suggests that STEP has a greater capacity for long-term learning and task improvement, particularly in more complex scenarios, where CLIN's performance tends to stall.

\subsubsection{The importance of Planner}
\paragraph{Ablation experiment. } To evaluate the necessity of the \textit{Planner} component, we conducted ablation experiments on STEP by removing the \textit{Planner} and retaining only the \textit{Executor}, \textit{Evaluator}, and \textit{Memory} modules (see Figure \ref{fig:framework}). In this configuration, the \textit{Executor} directly receives tasks from the environment and retrieves information from \textit{Memory} without any distillation or task refinement. Moreover, the absence of the \textit{Planner} eliminates the refinement of subtasks typically performed by the \textit{Evaluator}. 

\begin{figure*}[h]
    \centering
    \begin{subfigure}[b]{0.49\textwidth}
        \includegraphics[width=\textwidth]{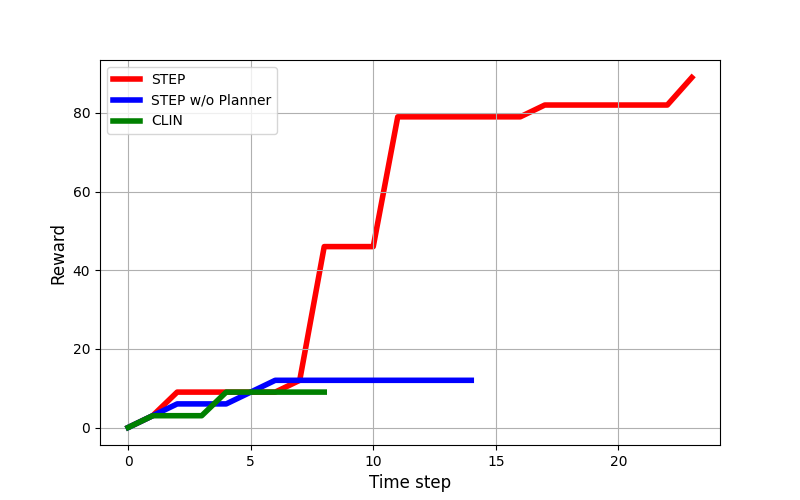}
        \caption{}
        \label{fig:3model}
    \end{subfigure}
    \hfill
    \begin{subfigure}[b]{0.49\textwidth}
        \includegraphics[width=\textwidth]{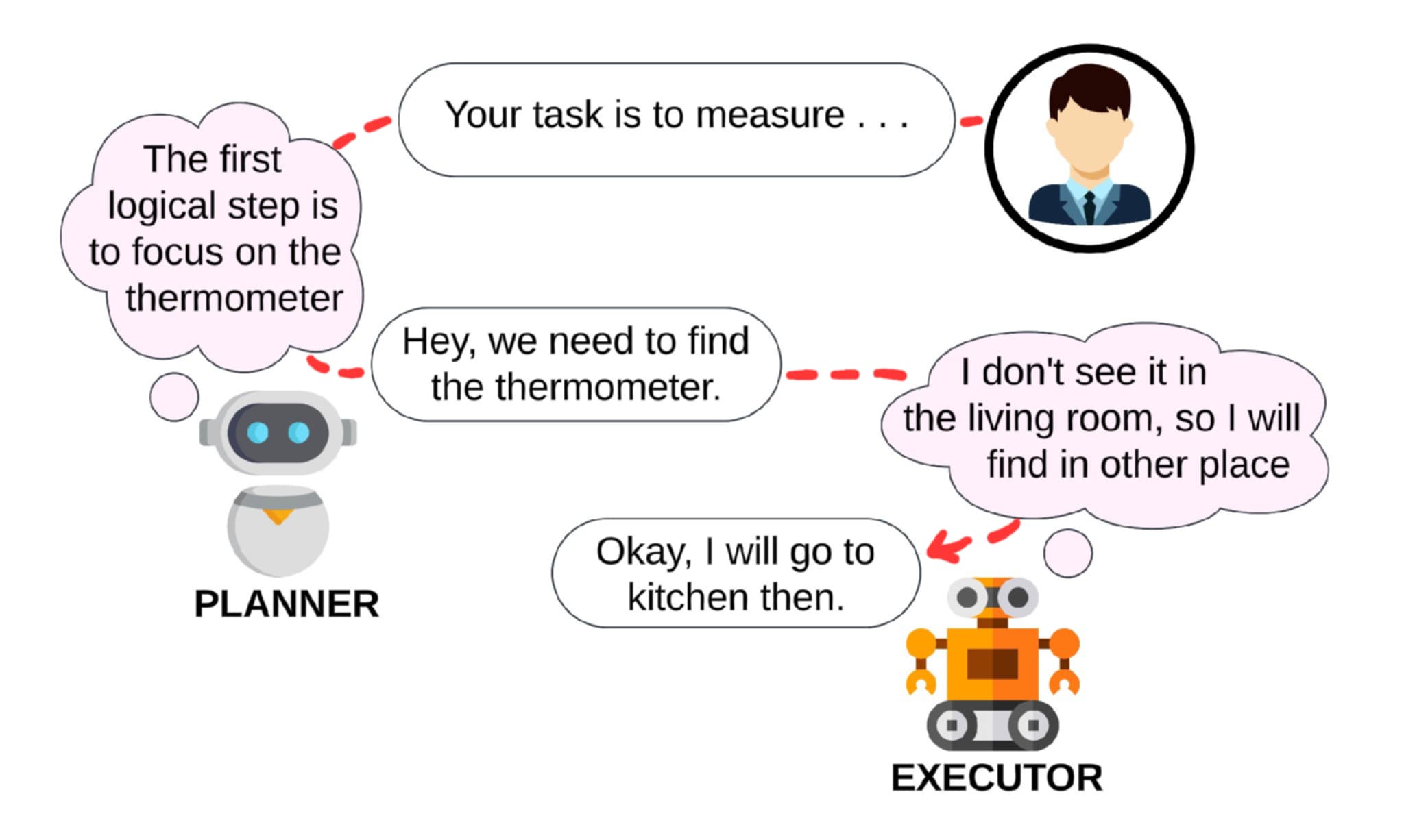}
        \caption{}
        \label{fig:interact}
    \end{subfigure}
    \caption{\textbf{(a)} Comparison of STEP, STEP w/o Planner and CLIN in $\text{Temp}_{1}$. \textbf{(b)} \textit{Planner} and \textit{Executor} interaction.}
    \label{fig:combined}
\end{figure*}

\begin{table}[ht]
\centering
\resizebox{0.8\columnwidth}{!}{%
\begin{tabular}{l||cc}
\toprule
\multicolumn{1}{l||}{} & \multicolumn{2}{c}{\textbf{STEP w/o Planner}} \\
\textbf{Task} & \textbf{$\Delta$avg score $(\downarrow)$} & \textbf{\%succ. rate $(\downarrow)$} \\
\midrule
\textbf{S} & 10.9 & 11.1 \\
\textbf{L} & 11.0 & 22.2 \\
\midrule
\textbf{All} & 10.9 & 16.7 \\
\bottomrule
\end{tabular}%
}
\caption{Average performance reduction without the Planner on short (S), long (L), and combined (All) tasks.}
\label{tab:ablation}
\end{table}

As presented in Table \ref{tab:ablation}, the results demonstrate a significant drop in performance. Specifically, the removal of the \textit{Planner} leads to a reduction in the average score by 10.9 points in short tasks and 11.0 points in long tasks. Furthermore, the success rate decreases by 11.1\% for short tasks and 22.2\% for long tasks, culminating in an overall decline of 10.9 points in average score and 16.7\% in success rate across all tasks. Notably, the average score of the STEP w/o Planner model is approximately the same as that of CLIN, with both achieving around 57 points overall. This result was expected, as the ablation model lacks the critical \textit{Planner} component, making it architecturally similar to CLIN. These findings also suggest that the addition of the \textit{Evaluator} (which CLIN does not include) does not significantly enhance performance in the absence of the \textit{Planner}.

\paragraph{A qualitative example. } Figure \ref{fig:3model} illustrates the optimal strategies employed by STEP, STEP w/o Planner, and CLIN during Task $\text{Temp}_{1}$, where the agent must measure the temperature of an unknown substance B. Although the location of substance B is known (in the living room), the agent must first locate a hidden thermometer. Without a planner, both CLIN and STEP w/o Planner instinctively head to the living room to interact with substance B, resulting in a -100 point penalty for skipping the necessary step of finding the thermometer. As reflected in the graph, this mistake leads to early termination for both models—CLIN is cut off at step 8 and STEP w/o Planner at step 14—with their rewards stagnating below 15 points due to incorrect task execution.

In contrast, STEP with \textit{Planner} adheres to the correct task order. The \textit{Planner} isolates the subtask of finding the thermometer, ensuring the \textit{Executor} focuses solely on the immediate task without being distracted by the known location of substance B (see Figure \ref{fig:interact}). This isolation prevents the \textit{Executor} from prematurely interacting with substance B and helps STEP avoid penalties. As reflected in the graph, STEP continues to progress smoothly, with its reward rapidly increasing and reaching a maximum of 90. 

Additionally, we observed that without the \textit{Planner}, the performance of the other models does not improve, even when \textit{Memory} correctly suggests the location of the thermometer (in the kitchen). The situation worsens when the models receive hints related to substance B, such as ``going to the living room is necessary to find substance B'' or ``focusing on substance B may contribute to the task.'' Without the high-level task distillation and guidance provided by the \textit{Planner}, these models are unable to effectively leverage the usefulness of information. This further underscores the \textit{Planner}’s critical role in task decomposition and knowledge distillation, ensuring the agent follows the task order.

\section{Limitation}

\paragraph{Poor subtask generation. } The agent can easily deviate from the intended task if the initial subtask goes off track. The \textit{Executor} relies entirely on the \textit{Planner}'s guidance, while the \textit{Planner} depends on the \textit{Executor}’s exploration to understand the environment. This interdependency can lead to significant problems if the \textit{Planner} generates an irrelevant subtask. In task Biology, the agent is required to \textit{“focus on the 3 life stages of the wolf”}.  Instead of guiding the agent to find a wolf outside the house, the \textit{Planner} incorrectly decides that painting a representation of a wolf is a valid solution: \textit{“Since I am in an art studio, I can use the paint to create representations of the wolf's life stages.”} The \textit{Planner} then directs the \textit{Executor} to gather materials, mix colors, and attempt to draw the wolf. 

Once the \textit{Planner} locks onto an incorrect high-level subtask, it becomes increasingly difficult for the agent to revise its approach. As a result, it falls into a repetitive loop of failure, with the \textit{Executor} never leaving the house to discover the real wolf. The only potential way to break this loop is through a refinement mechanism; however, in this case, the mechanism fails to detect the irrelevance of the subtask and continues allowing the agent to explore the painting activity. Instead of exploring horizontally, the agent dives vertically into the wrong path.
\paragraph{Poor strategy generation.} After each trial, a strategy is generated to inform future agents of the latest successful traces. Whether provided with a list of completed subtasks or a full action trace, the model struggles to eliminate unnecessary elements. For instance, in the Force task, where the agent must \textit{``determine which of the two inclined planes has the steepest angle''}, the agent first needs to locate the inclined planes in the workshop. After extensive exploration—navigating to the art studio, hallway, outside, and then back to the hallway—the agent finally finds the workshop and continues with the task. However, at the end of the episode, the agent fails to eliminate redundant looping traces, such as repeatedly moving to the hallway without gaining useful information, and instead groups these actions into the vague category of\textit{``Exploring the environment.''}

\textbf{Balancing strategy abstraction presents significant challenges.} Strategies that are too detailed closely resemble action traces, causing future agents to simply replicate past actions without considering variations in context. In contrast, overly abstract strategies result in unachievable goals, as they lack the necessary steps for effective task completion. To mitigate this, our approach attempts to retain actions that gain rewards from the environment, then prompt LLMs to identify which remaining actions are essential, before abstracting the new action trace. However, this method is still far from optimal and highlights a gap for future research.

\section{Conclusion}
We present STEP, a simple yet effective framework that leverages LLM as a stepwise planner. STEP enhances memory utilization and task sequence recognition, achieving state-of-the-art performance in the ScienceWorld benchmark. By showcasing the potential of STEP, our research contributes to the ongoing exploration and development of language agents as planners.

\bibliography{main}

\appendix
\section{ScienceWorld Tasks Action Space}\label{app:actionspace}
The ScienceWorld action space is reported in Table~\ref{actionspace}.
\begin{table}[htbp]
\centering
\resizebox{0.5\textwidth}{!}{%
\begin{tabular}{ll}
\toprule
\textbf{Action} & \textbf{Description} \\ 
\midrule
open/close OBJ & open/close a container \\
de/activate OBJ & activate/deactivate a device \\
connect OBJ to OBJ & connect electrical components \\
disconnect OBJ & disconnect electrical components \\
use OBJ [on OBJ] & use a device/item \\
look around & describe the current room \\
look at OBJ & describe an object in detail \\
look in OBJ & describe a container’s contents \\
read OBJ & read a note or book \\
move OBJ to OBJ & move an object to a container \\
pick up OBJ & move an object to the inventory \\
put down OBJ & drop an inventory item \\
pour OBJ into OBJ & pour a liquid into a container \\
dunk OBJ into OBJ & dunk a container into a liquid \\
mix OBJ & chemically mix a container \\
go to LOC & move to a new location \\
teleport to LOC * & teleport to a specific room \\
eat OBJ & eat a food \\
flush OBJ & flush a toilet \\
focus on OBJ & signal intent on a task object \\
wait [DURATION] & take no action for some duration \\
task & describe current task \\
inventory & list agent’s inventory \\
\bottomrule
\end{tabular}
}
\caption{The 25 actions in the action space of \textsc{Science-World}. Actions can take up to two parameters, referencing objects the action should interact with. * signifies that the \textit{teleport} action is only available to agents in a simplified mode.}\label{actionspace}
\end{table}

\section{Example of STEP trajectory for a task}\label{app:trajectory}
We present a sample of task trajectory for STEP in Figure~\ref{fig:task-trajectory}.

\begin{figure*}[t]

\begin{framed}
    \fontsize{10}{12}\selectfont
    \textbf{Task Description:} Your task is to measure the melting point of lead, which is located around the kitchen. First, focus on the thermometer. Next, focus on the lead. If the melting point of lead is above 50.0 degrees celsius, focus on the red box. If the melting point of lead is below 50.0 degrees celsius, focus on the green box. The boxes are located around the kitchen.\\

    \lstset{basicstyle=\ttfamily\fontsize{9}{7}\selectfont, breaklines=true}
    \begin{lstlisting}
    > look around
    This room is called the art studio. In it, you see: 
    	the agent
    	a substance called air
    	a large cupboard. The large cupboard door is closed. 
    	a table. On the table is: a bowl (containing nothing).
    	a wood cup (containing blue paint)
    	a wood cup (containing red paint)
    	a wood cup (containing yellow paint)
    You also see:
    	A door to the hallway (that is open)
    
    > go hallway
    You move to the hallway.
    
    > go greenhouse
    You move to the greenhouse.
    
    > look around
    This room is called the greenhouse. In it, you see: 
    	the agent
    	a substance called air
    	a bee hive. The bee hive door is closed. 
    	a jug (containing nothing)
    	a shovel
    	a sink, which is turned off. In the sink is: nothing.
    You also see:
    	A door to the hallway (that is open)
    	A door to the outside (that is open)
    
    > go outside
    You move to the outside.
    
    > go kitchen
    You move to the kitchen.
    
    > focus on thermometer
    You focus on the thermometer.
    
    > pick up cup containing lead
    You move the plastic cup to the inventory.
    
    > focus on lead
    You focus on the lead.
    
    > focus on thermometer
    You focus on the thermometer.
    
    > look at thermometer
    a thermometer, currently reading a temperature of 10 degrees celsius
    
    > look around
    This room is called the kitchen. In it, you see: 
    	the agent
    	a substance called air
    	a chair. On the chair is: nothing.
            ...
    	a table. On the table is: a glass cup (containing nothing).
    	a thermometer, currently reading a temperature of 10 degrees celsius
    You also see:
    	A door to the bathroom (that is open)
    	A door to the hallway (that is open)
    	A door to the outside (that is open)
    
    > focus on lead
    You focus on the lead.
    
    > focus on red box
    You focus on the red box. (Task Completed!)
    \end{lstlisting}
\end{framed}
\caption{Example of task $\text{Temp}_{2}$ trajectory with STEP.}\label{fig:task-trajectory}
\end{figure*}

\end{document}